\documentclass{article}

\usepackage{microtype}
\usepackage{graphicx}
\usepackage{subcaption}
\usepackage{booktabs} 

\usepackage{hyperref}

\usepackage[preprint]{icml2026}

\usepackage{amsmath}
\usepackage{amssymb}
\usepackage{mathtools}
\usepackage{amsthm}
\usepackage{multirow}
\usepackage{bm}
\usepackage{booktabs}
\usepackage[table]{xcolor} 
\definecolor{lightgray}{gray}{0.92}
\usepackage[capitalize,noabbrev]{cleveref}

\theoremstyle{plain}

\theoremstyle{definition}

\theoremstyle{remark}

\usepackage[textsize=tiny]{todonotes}
\usepackage{cuted} 
\icmltitlerunning{LoPA: Scaling dLLM Inference via Lookahead Parallel Decoding}

\begin{document}

\twocolumn[
  \icmltitle{LoPA: Scaling dLLM Inference via Lookahead Parallel Decoding}



  \icmlsetsymbol{equal}{*}
  \icmlsetsymbol{corr}{\textdagger}  
  \begin{icmlauthorlist}
    \icmlauthor{Chenkai Xu}{sjtu,equal}
    \icmlauthor{Yijie Jin}{sjtu,equal}
    \icmlauthor{Jiajun Li}{huawei}
    \icmlauthor{Yi Tu}{huawei}
    \icmlauthor{Guoping Long}{huawei}
    \icmlauthor{Dandan Tu}{huawei}
    \icmlauthor{Mingcong Song}{huawei}
    \icmlauthor{Hongjie Si}{huawei}
    \icmlauthor{Tianqi Hou}{huawei}
    \icmlauthor{Junchi Yan}{sjtu}
    \icmlauthor{Zhijie Deng}{sjtu,corr}

    \vspace{0.1cm} 
    \centering 
    Github: \url{https://github.com/zhijie-group/LoPA} \\ 
  \end{icmlauthorlist}

  \icmlaffiliation{sjtu}{Shanghai Jiao Tong University}
  \icmlaffiliation{huawei}{Huawei}

  \icmlcorrespondingauthor{Zhijie Deng}{zhijied@sjtu.edu.cn}

  \icmlkeywords{Machine Learning, ICML}

]



\printAffiliationsAndNotice{}  

\begin{strip}
    \centering
    \includegraphics[width=0.9\textwidth]{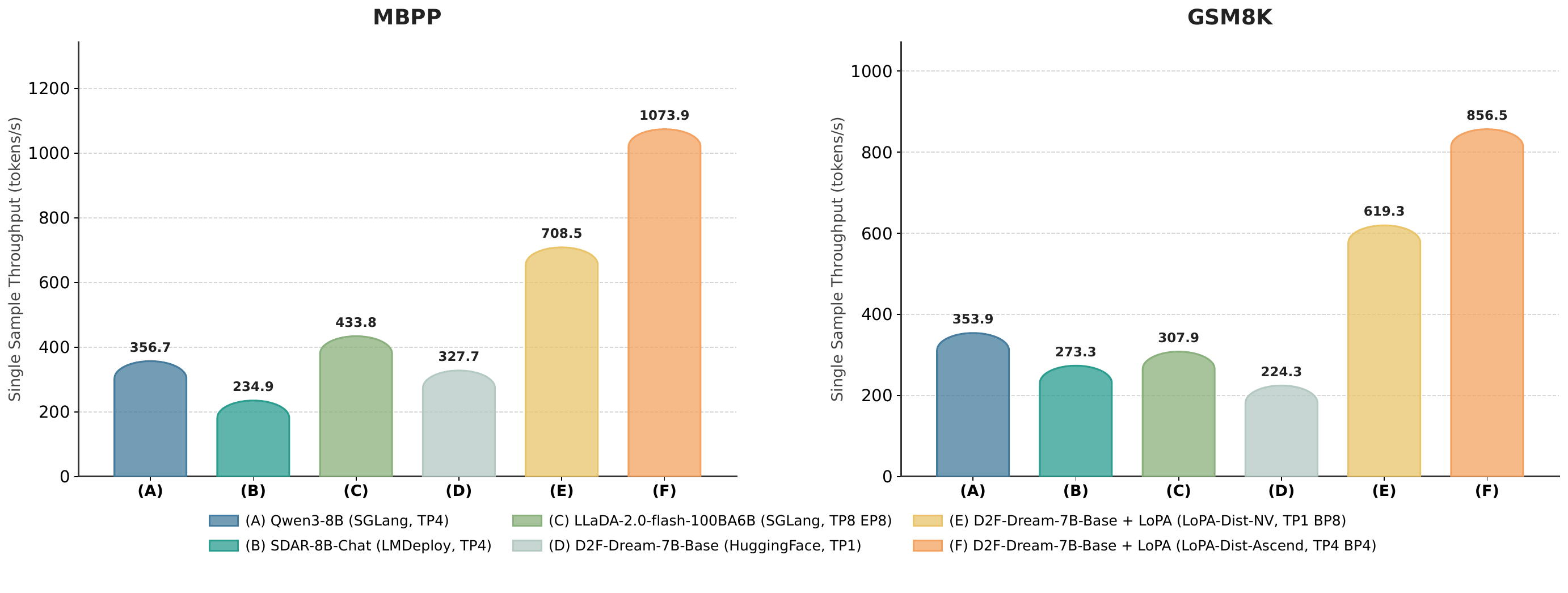} 
    \captionof{figure}{\textbf{Throughput performance of LoPA.} LoPA accelerates the \textbf{single-sample throughput} for D2F-Dream to up to 1073.9 and 856.5 tokens/s on MBPP and GSM8K respectively, significantly outperforming baselines. More details are provided in Table~\ref{tab:baseline_results}.} 
    \label{fig:tps_comparison}
\end{strip}

\begin{abstract} 
Diffusion Large Language Models (dLLMs) have demonstrated significant potential for high-speed inference. 
However, current confidence-driven decoding strategies are constrained by limited parallelism, typically achieving only 1--3 tokens per forward pass (TPF). 
In this work, we identify that the degree of parallelism during dLLM inference is highly sensitive to the Token Filling Order (TFO). Then, we introduce \textbf{Lo}okahead \textbf{PA}rallel Decoding (\textbf{LoPA}), a training-free, plug-and-play algorithm, to identify a superior TFO and hence accelerate inference. LoPA concurrently explores distinct candidate TFOs via parallel branches, and selects the one with the highest potential for future parallelism based on branch confidence. We apply LoPA to the state-of-the-art D2F model and observe a substantial enhancement in decoding efficiency. Notably, LoPA increases the TPF of D2F-Dream to 10.1 on the GSM8K  while maintaining performance superior to the Dream baseline. Furthermore, to facilitate this unprecedented degree of parallelism, we develop a specialized multi-device inference system featuring Branch Parallelism (BP), which achieves a single-sample throughput of \textbf{1073.9} tokens per second under multi-GPU deployment. 
\end{abstract}

\section{Introduction}

\begin{figure*}[t]
    \centering
    \includegraphics[width=0.9\textwidth]{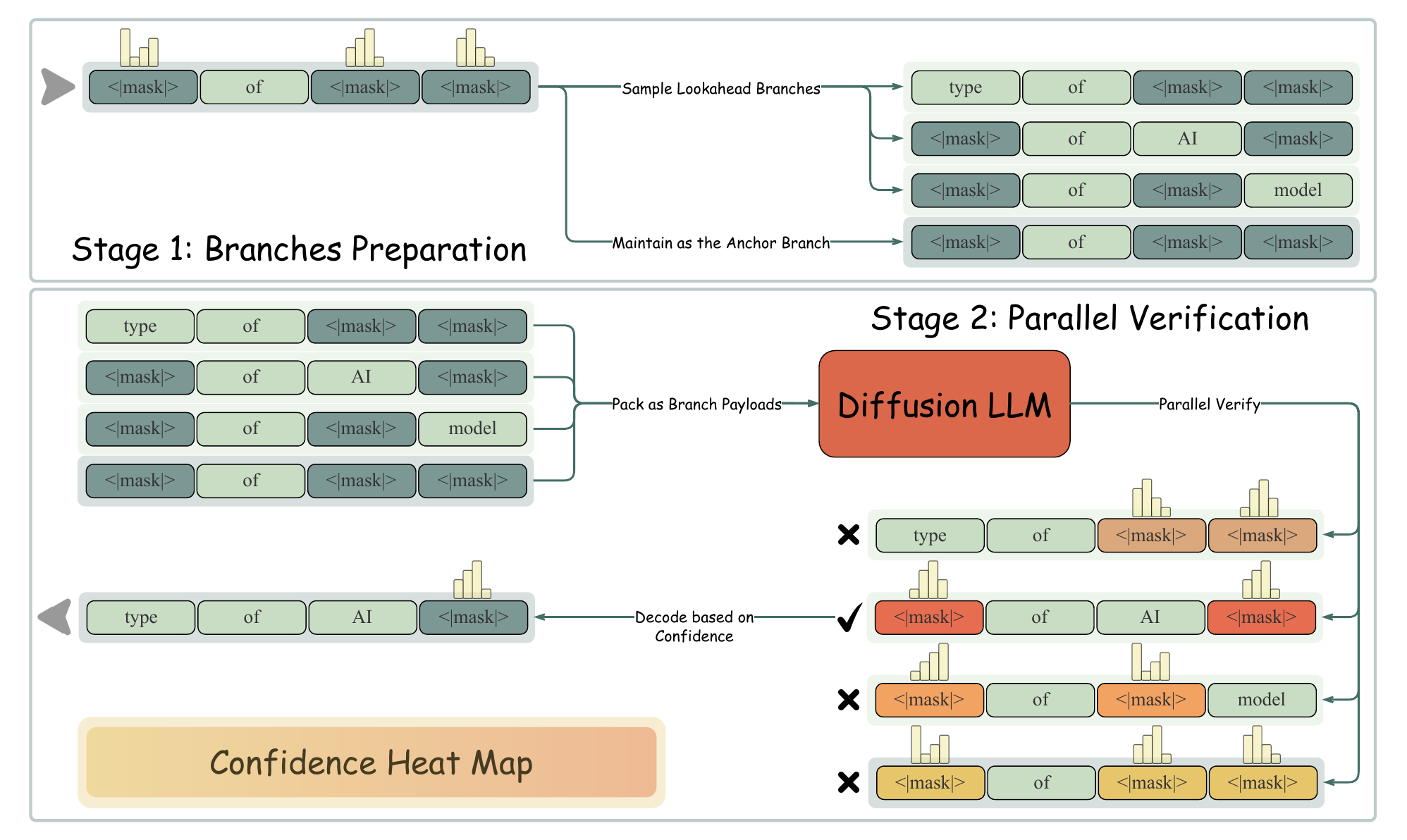} 
    \caption{\textbf{Overview of Lookahead Parallel Decoding (LoPA)}. In each iteration, LoPA generates a anchor branch alongside multiple lookahead branches (e.g., $B_1, \dots, B_k$) by independently sampling high-confidence positions from the baseline's unfilled set. A branch confidence verification mechanism then evaluates all branches in parallel within a single forward pass, selecting the optimal path to maximize future parallelism.}
    \label{fig:lopa_overview}
\end{figure*}

Diffusion Large Language Models (dLLMs)~\cite{llada, dream, diffucoder, d2f, sdar} have emerged as a highly promising paradigm for text generation. By iteratively refining a full-mask sequence into text tokens, dLLMs decouple generation depth from sequence length, theoretically offering superior potential for inference speed. Recent studies~\cite{d2f, sdar} have demonstrated speed surpassing autoregressive (AR) models, thereby providing higher throughput for latency-sensitive applications.

Despite this potential, the practical parallelism of state-of-the-art dLLMs remains constrained. Leading models such as Fast-dLLM~\cite{fast-dllm}, D2F~\cite{d2f}, and SDAR~\cite{sdar} employ confidence-driven sampling, which fills tokens exceeding a confidence threshold $\tau$ in each iteration. However, this strategy typically yields only 1--3 tokens per forward pass (TPF) on common tasks such as mathematics and coding, failing to fully unleash the parallel potential of dLLMs.

Our investigation traces this limitation to a fundamental sensitivity: parallelism is bounded by prediction confidence and the confidence is heavily influenced by the Token Filling Order (TFO). As observed in LLaDA~\cite{llada}, varying TFOs significantly shifts the generative distribution and confidence landscape. Consequently, the standard strategy of greedily prioritizing positions with the highest current confidence may lead to suboptimal trajectories, raising the question: \textit{Can we actively explore superior TFOs to maximize future confidence and unlock higher parallelism?}

To this end, we introduce \textbf{Lo}okahead \textbf{PA}rallel Decoding (\textbf{LoPA}), a training-free, plug-and-play algorithm designed to search for TFOs with high future parallelization potential. LoPA operates in three phases per iteration: (1) advancing decoding by sampling an anchor branch $B_0$ via standard confidence-driven strategies; (2) exploring distinct TFOs beyond $B_0$ by generating $k$ lookahead branches, each sampling from the top-$k$ high-confidence candidate positions to ensure reliable exploration coverage~\cite{fast-dllm}; and (3) identifying the optimal path by verifying all $k+1$ branches in a single forward pass to retain the one with the highest future parallelization potential. By iteratively selecting optimal branches, LoPA can boost overall TPF.

We integrate LoPA with D2F~\cite{d2f}, scaling the TPF of D2F-Dream to 10.1 on GSM8K~\cite{gsm8k} while maintaining performance scores surpassing the original Dream baseline, and scaling D2F-DiffuCoder to 8.3 on HumanEval+~\cite{humaneval,evalplus} with marginal performance degradation. To support LoPA, we co-design a multi-device inference system that distributes branches across devices on multiple platforms, achieving a single-sample throughput of \textbf{1073.86} tokens/s. We further validate LoPA's generalizability by integrating it with Vanilla Dream~\cite{dream}. Evaluations demonstrate that LoPA effectively scales dLLM parallelism, establishing a clear, controllable speed-accuracy trade-off.

Our contributions are summarized as follows:
\begin{itemize}
    \item We identify TFO as a key factor influencing dLLM parallelism and propose LoPA, a training-free algorithm that looks ahead to optimize TFO.
    \item We demonstrate that LoPA scales the TPF of D2F-Dream to 10.1 on GSM8K and D2F-DiffuCoder to 8.3 on HumanEval+, while maintaining comparable performance.
    \item We develop a specialized Branch Parallel inference system, achieving near-linear scalability and a throughput of \textbf{1073.86} tokens/s.
\end{itemize}

\section{Related Work}
\label{sec:related_work}

\textbf{Diffusion Large Language Models (dLLMs).} Autoregressive (AR) models~\cite{gpt4, llama, mistral, deepseek} have long dominated text generation, yet their sequential decoding imposes an inherent latency bottleneck. To address this, dLLMs~\cite{llada, dream, diffucoder, deepmind_dllm} have emerged as a non-autoregressive paradigm. By iteratively denoising fully masked sequences, dLLMs enable parallel token prediction and leverage bidirectional attention for holistic context modeling. Recent scaling efforts, whether training from scratch~\cite{llada} or initializing from pre-trained AR weights~\cite{dream}, have yielded dLLMs with performance competitive to state-of-the-art AR models, validating their potential for high-quality, parallel generation.

\textbf{Acceleration of dLLMs.} Despite their parallel nature, dLLM inference remains computationally expensive due to multi-step denoising and incompatibility with standard KV caching. Acceleration strategies fall into two primary paradigms. Training-based methods compress sampling steps or restructure generation; for instance, dParallel~\cite{dparallel} applies consistency distillation, while D2F~\cite{d2f} employs asymmetric distillation to enable block-autoregressive pipelining. Training-free methods optimize inference without weight updates. One avenue adapts KV caching to bidirectional attention via approximate schemes~\cite{dllm_cache, fast-dllm, dkv_cache}. Another focuses on heuristic decoding optimizations, where works like Fast-dLLM~\cite{fast-dllm}, Prophet~\cite{prophet}, and Credit Decoding~\cite{credit_decoding} exploit confidence patterns or early-layer determinism to skip redundant steps.

\textbf{Speculative Decoding.} Speculative decoding, a standard for AR acceleration~\cite{leviathan2023, chen2023, spector2023}, employs efficient draft models for parallel verification. Innovations include tree-based drafting~\cite{miao2024, eagle}, multi-head structures~\cite{medusa}, and draft-free fixed-point iterations~\cite{lookahead_ar, zhang2024draft}. In the dLLM domain, speculative concepts have evolved from relying on external AR guidance~\cite{adaptive_parallel} to self-verification methods like Spiffy~\cite{spiffy} and Free Draft-and-Verification~\cite{free_draft}. While the latter achieve lossless acceleration by maximizing token acceptance within a fixed generation distribution, our approach fundamentally diverges. Instead of passively verifying a static sequence, LoPA actively explores TFO to discover trajectories with superior future confidence, effectively optimizing the output distribution to unlock parallelism beyond standard greedy limits.

\section{Methodology}

This section first explains the foundational Confidence-Driven Sampling used in regular dLLM inference~\cite{fast-dllm} and then elaborates on LoPA.

\subsection{Preliminary: Confidence-Driven Sampling for dLLMs}
Confidence-driven sampling is a prevalent paradigm for current dLLMs to boost parallelism, which has been widely adopted in advanced methods such as Fast-dLLM~\cite{fast-dllm}, D2F~\cite{d2f}, and SDAR~\cite{sdar}. Specifically, given a sequence $x_t$ with a set of masked positions $M_t$, the dLLM model $p_\theta$ outputs a predictive distribution $p_\theta(\cdot | x_t)$. A candidate sequence $\hat{x}_0 \sim p_\theta(\cdot | x_t)$ is sampled, and a confidence function, $\text{Conf}(\cdot)$, assigns a score to each position $i \in M_t$. The set of positions to fill, $I_{fill}$, is then determined as:

\begin{equation}
    \label{eq:ifill}
    \begin{split}
    S_{\text{high}} &= \{i \in M_t \mid \text{Conf}(i) > \tau\} \\
    I_{fill} &= \begin{cases} 
        S_{\text{high}} & \text{if } S_{\text{high}} \neq \emptyset \\ 
        \{\arg\max_{i \in M_t} \text{Conf}(i)\} & \text{otherwise} 
    \end{cases}
    \end{split}
\end{equation}

The algorithm then accepts the predictions according to $I_{fill}$ and moves to the next iteration.

\subsection{Lookahead PArallel Decoding
(LoPA)}

\begin{algorithm}[tb]
   \caption{Lookahead Parallel Decoding (LoPA)}
   \label{alg:lopa}
   \small
\begin{algorithmic}
   \STATE {\bfseries Input:} Sequence $x_t$, Mask $M_t$, Branch budget $k$
   \STATE {\bfseries Output:} Updated Sequence $x_{t+1}$, Mask $M_{t+1}$
   
   \STATE \textbf{// 1. Anchor Branch Construction}
   \STATE Compute distribution $p_\theta(\cdot|x_t)$ and confidence scores
   \STATE Determine anchor fill set $I_{fill}$ via \textbf{Eq.~\ref{eq:ifill}} to form branch $B_0$
   \STATE Identify remaining unfilled set $M_{B_0} \leftarrow M_t \setminus I_{fill}$
   
   \STATE \textbf{// 2. Lookahead Branches Spawning}
   \STATE Select top-$k$ positions $\{p_1, \dots, p_k\} \subset M_{B_0}$ with highest confidence
   \FOR{$j=1$ {\bfseries to} $k$}
       \STATE Construct branch $B_j$ by independently sampling position $p_j$
   \ENDFOR
   
   \STATE \textbf{// 3. Parallel Verification}
   \STATE Concatenate $\{B_0, \dots, B_k\}$ as a batch
   \STATE Compute scores $C(B_j)$ via \textbf{Eq.~\ref{eq:branch_conf}} (Single Pass)
   \STATE Select optimal branch $B^* \leftarrow \arg\max_{j} C(B_j)$
   
   \STATE {\bfseries Return} Update $x_{t+1}, M_{t+1}$ according to $B^*$
\end{algorithmic}
\end{algorithm}

\begin{figure*}[t]
    \centering
    \includegraphics[width=0.9\textwidth]{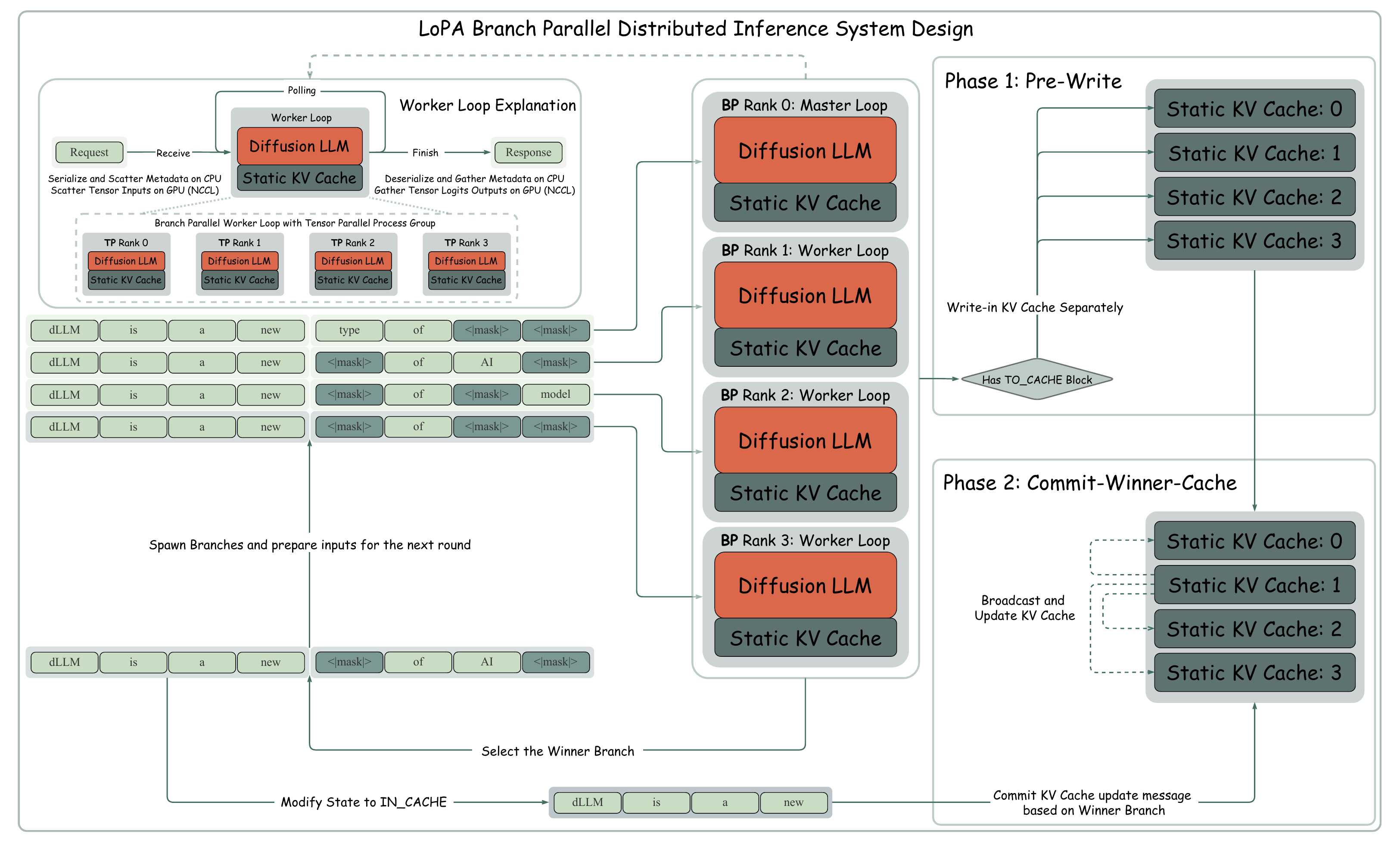} 
    \caption{\textbf{Overview of LoPA Branch Parallel Distributed Inference System Design.} A key distinction lies in the KV cache management protocol tailored for different backends: \textbf{LoPA-Dist-NV} utilizes a robust two-phase update mechanism to ensure consistency, whereas \textbf{LoPA-Dist-Ascend} adopts a streamlined single-phase update strategy for optimized serving efficiency.}
    \label{fig:lopa_sys}
\end{figure*}
As shown in Figure~\ref{fig:lopa_overview}, LoPA addresses the limitation of fixed sampling by looking ahead at multiple TFOs in every decoding iteration. It generates multiple sampling branches concurrently and identifies the one with superior potential for parallel decoding. The detailed procedure is outlined in Algorithm~\ref{alg:lopa}.

\paragraph{Look ahead Multiple TFOs in Parallel.}
Standard confidence-driven sampling relies on a single anchor branch where only positions in $I_{fill}$ are sampled. LoPA extends this by exploring one step further. To ensure exploration is both effective and reliable, 
we prioritize sampling tokens with higher confidence, a strategy that has been proved in Fast-dLLM ~\cite{fast-dllm} to yield more stable predictions. 
Specifically, in addition to the anchor branch $B_0$, we generate $k$ competitive branches. We identify the top-$k$ positions from the unfilled set $M_{B_0}$ that possess the highest confidence scores. For each identified position, we sample it independently to create a distinct branch. This results in a set of $k$ new branches $\{B_1, \dots, B_k\}$, each possessing a unique partially filled sequence $x_{B_j}$ and a unfilled set $M_{B_j}$.

\paragraph{Branch Confidence-based Verification.}
Inspired by DeepConf~\cite{deepthinkconfidence}, we design a branch confidence metric to guide the selection among candidate decoding paths. Formally, the confidence of a branch $B_j$ is defined as the average prediction confidence over its remaining unfilled positions $M_{B_j}$:
\begin{equation}
\label{eq:branch_conf}
C(B_j) = \frac{1}{|M_{B_j}|} \sum_{i \in M_{B_j}} \text{Conf}(i)
\end{equation}
A higher branch confidence indicates that more unfilled positions are likely to be accepted in the next decoding step. This directly increases the number of tokens filled per iteration, thereby enhancing the overall parallelism.
Beyond this mean confidence, branch confidence can also be quantified by other methods~\cite{deepthinkconfidence}, such as applying a sliding window to assess local quality or averaging confidence over the least confident segment to identify weak links. 
We adopt this naive average confidence as the default for its simplicity and robust performance.

This branch confidence verification mechanism offers distinct advantages. First, it facilitates the packing and verification of all candidate branches within a single forward pass. Second, the logits computed during evaluation are repurposed for the subsequent decoding step, obviating the need for additional computation.

\begin{figure*}[t]
    \centering
    \includegraphics[width=0.8\textwidth]{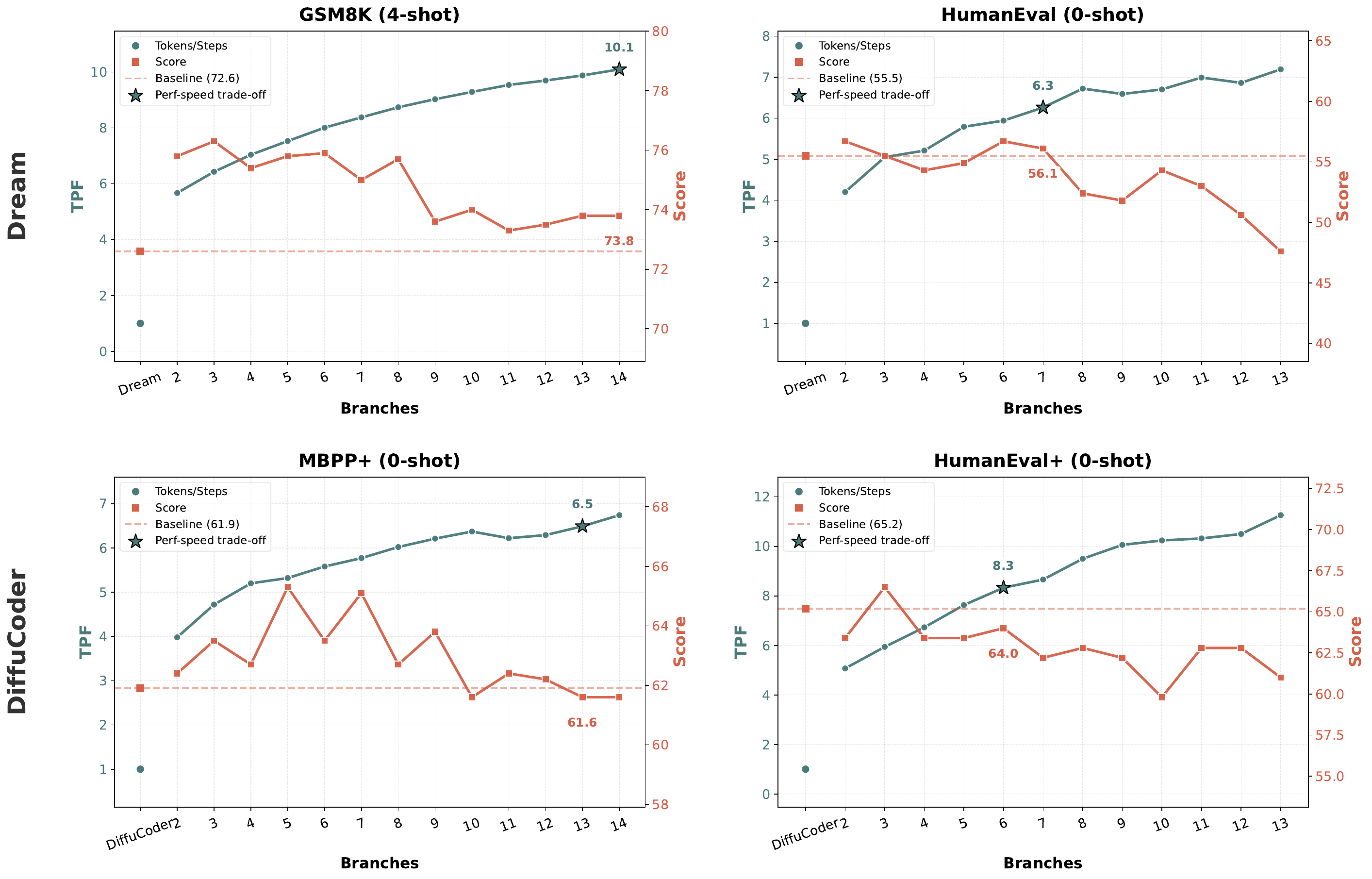} 
    \caption{\textbf{Scaling Curves of LoPA.} LoPA scales the tokens per forward pass (TPS) for D2F-Dream and D2F-DiffuCoder to up to 10.1 and 8.3 on GSM8k and HumanEval+ respectively, with comparable
    performance.}
    \label{fig:scale}
\end{figure*}

\subsection{Application: Integration with D2F}
LoPA integrates seamlessly with D2F~\cite{d2f} by treating all active blocks as a single window for branch exploration. Crucially, within this window, we replace the original block-level causal attention with full attention.

This design yields two key benefits. First, the simplified attention mechanism significantly reduces complexity and enhances compatibility with mainstream inference frameworks, directly boosting computational speed. Second, we empirically find that this localized full attention maintains or even improves generation quality. By enabling blocks to attend to ``future'' tokens within the limited window, the information flow is enriched without disrupting the global causal dependency required for valid generation.

\begin{table*}[t]
    \centering
    \setlength{\tabcolsep}{5pt}
    \resizebox{0.95\textwidth}{!}{%
    \begin{tabular}{llcccccccc}
        \toprule
        \multirow{2}{*}{\textbf{Model}} & \multirow{2}{*}{\textbf{Decoding Algo}} & \multicolumn{2}{c}{\textbf{MBPP 3-shot}} & \multicolumn{2}{c}{\textbf{Math 4-shot}} & \multicolumn{2}{c}{\textbf{HumanEval 0-shot}} & \multicolumn{2}{c}{\textbf{GSM8K 4-shot}} \\
        \cmidrule(lr){3-4} \cmidrule(lr){5-6} \cmidrule(lr){7-8} \cmidrule(lr){9-10}
         & & \textbf{TPF} & \textbf{Score} & \textbf{TPF} & \textbf{Score} & \textbf{TPF} & \textbf{Score} & \textbf{TPF} & \textbf{Score} \\
        \midrule
        
         & Vanilla & 1.0 & \textbf{56.2} & 1.0 & 33.7 & 1.0 & 55.5 & 1.0 & 72.6 \\
         & Fast-dLLM & 1.9 & 55.6 & 1.9 & \textbf{37.6} & 1.8 & 55.5 & 2.1 & 72.6 \\
        \rowcolor{lightgray} \cellcolor{white} \multirow{-3}{*}{Dream} & LoPA & 3.3 & 54.8 & 3.4 & 37.0 & 2.9 & 53.0 & 3.1 & 73.3 \\
        \midrule
        
         & Vanilla & 2.3 & 53.8 & 2.6 & 36.8 & 2.5 & \textbf{56.1} & 3.1 & \textbf{78.5} \\
        \rowcolor{lightgray} \cellcolor{white} \multirow{-2}{*}{D2F-Dream} & {LoPA} & \textbf{5.4} & 56.0 & \textbf{8.0} & 35.2 & \textbf{6.3} & \textbf{56.1} & \textbf{10.1} & 73.8 \\
        \bottomrule
    \end{tabular}
    }
    \caption{\textbf{Accuracy-preserving parallelism scaling of Dream on multiple benchmarks across multiple branches.} TPF denotes Tokens Per Forward pass. LoPA significantly scales the TPF of D2F-Dream while maintaining or exceeding baseline scores.}
    \label{tab:dream_performance}
\end{table*}

\begin{table*}[t]
    \centering
    \begin{tabular}{llcccc}
        \toprule
        \multirow{2}{*}{\textbf{Model}} & \multirow{2}{*}{\textbf{Decoding Algo}} & \multicolumn{2}{c}{\textbf{MBPP+}} & \multicolumn{2}{c}{\textbf{HumanEval+}} \\
        \cmidrule(lr){3-4} \cmidrule(lr){5-6}
         & & \textbf{TPF} & \textbf{Score} & \textbf{TPF} & \textbf{Score} \\
        \midrule
        
        DiffuCoder & Vanilla & 1.0 & 61.9 & 1.0 & 65.2 \\
        \midrule
        
        
        \multirow{2}{*}{D2F-DiffuCoder} & Vanilla & 2.2 & 61.9 & 2.2 & 65.9 \\
        
         & \cellcolor{lightgray}{LoPA} & \cellcolor{lightgray}\textbf{6.7} & \cellcolor{lightgray}61.6 & \cellcolor{lightgray}\textbf{8.3} & \cellcolor{lightgray}64.0 \\
        \bottomrule
    \end{tabular}
    \caption{\textbf{Accuracy-preserving parallelism scaling of DiffuCoder on MBPP+ and HumanEval+ benchmarks.} LoPA boosts TPF by nearly $4\times$ compared to the vanilla D2F baseline with minimal impact on generation quality.}
    \label{tab:diffucoder_performance}
\end{table*} 

\subsection{System Implementation}

To fully unleash the parallelism inherent in LoPA, we introduce \textbf{LoPA-Dist}, a high-throughput distributed inference system co-designed with the LoPA algorithm pipeline. LoPA-Dist introduces \textit{Branch Parallelism (BP)} to distribute candidate branches across multiple computing devices, orchestrating synchronized execution to maximize hardware utilization. We provide two specialized implementations of LoPA-Dist tailored for different hardware ecosystems and deployment scenarios.

\textbf{LoPA-Dist-NV: Latency-Oriented Optimization on CUDA.}
Targeting the NVIDIA CUDA platform, we developed \textbf{LoPA-Dist-NV}, a specialized implementation optimized for ultra-low latency single-sample acceleration. This system employs a pre-allocated static KV cache tightly coupled with the model architecture, effectively circumventing the overhead of dynamic object instantiation and the costly scattering of KV cache tensors via NCCL. To maintain context consistency across divergent branches without sacrificing speed, LoPA-Dist-NV implements a novel two-phase update protocol.

As illustrated in Figure~\ref{fig:lopa_sys}, during the forward pass, the system executes a \textit{Pre-Write} phase: each device speculatively writes the features of its \texttt{TO\_CACHE} blocks directly into designated cache slots. Since branch divergence leads to temporary cache inconsistency, we execute a \textit{Commit-Winner-Cache} phase: once the optimal branch is identified, its corresponding KV cache features are broadcast to all peer devices, overwriting local entries to enforce global synchronization. To further minimize runtime latency, LoPA-Dist-NV integrates FlashAttention~\cite{flash_attn} backends via SDPA, pre-merges LoRA weights, utilizes fused kernels for RMS Norm and RoPE, and implements an approximate prefix caching mechanism.

\textbf{LoPA-Dist-Ascend: High-Performance Serving on Ascend 910C.}
Building upon the foundation of LoPA-Dist-NV, we architected \textbf{LoPA-Dist-Ascend}, a high-throughput serving engine specifically optimized to exploit the computational power of the Huawei Ascend 910C. While retaining the core algorithmic logic of the CUDA implementation, LoPA-Dist-Ascend adopts a vLLM-like~\cite{vllm} architecture with a hybrid parallelism strategy orchestrated via process groups. Specifically, we assign four NPUs to each branch (TP4) to accelerate single-pass forward propagation through Tensor Parallelism, while scaling throughput via Branch Parallelism across groups.

To streamline system complexity for large-scale serving, LoPA-Dist-Ascend utilizes a block-wise causal mask for the attention mechanism. This design choice eliminates the necessity for the asynchronous \textit{Commit-Winner-Cache} phase; instead, it relies solely on a direct \textit{Pre-Write} operation, which inherently maintains inter-branch consistency through masking. To achieve optimal throughput, LoPA-Dist-Ascend integrates a suite of hardware-aware optimizations:

\begin{itemize}
\item \textit{NPU-Aware FlashAttention \& Pipelining:} Addressing the self-attention bottleneck, we implemented a custom FlashAttention~\cite{flash_attn} kernel optimized for the Ascend 910C's Cube Units. We further integrate advanced pipelining techniques~\cite{xy-serve} to effectively mask latencies between Cube and Vector operations. By tiling Q, K, and V matrices into the on-chip SRAM, we minimize HBM data movement, achieving near-theoretical IO throughput.

\item \textit{Graph Compilation \& Operator Fusion:} Since dLLMs involve iterative denoising steps necessitating repetitive kernel launches, we utilize graph compilation to fuse element-wise operations (e.g., bias addition and activations), significantly reducing launch overhead.

\item \textit{Diffusion-Tailored Memory Management:} We implement a paged memory management system similar to PagedAttention but specifically tailored for the state-space requirements of diffusion steps.

\item \textit{System-Level Optimizations:} The engine further incorporates W8A8 quantization, QKV merging, and fully asynchronous inference scheduling.

\end{itemize}

\begin{table*}[t]
    \centering
    \setlength{\tabcolsep}{4pt}
    \resizebox{0.95\textwidth}{!}{%
    \begin{tabular}{llcccccccc}
        \toprule
        \multirow{2}{*}{\textbf{Model}} & \multirow{2}{*}{\textbf{Platform}} & \multicolumn{4}{c}{\textbf{MBPP}} & \multicolumn{4}{c}{\textbf{GSM8K}} \\
        \cmidrule(lr){3-6} \cmidrule(lr){7-10}
         & & \textbf{Avg TPS} & \textbf{Max TPS} & \textbf{TPF} & \textbf{Latency} & \textbf{Avg TPS} & \textbf{Max TPS} & \textbf{TPF} & \textbf{Latency} \\
        \midrule
        
         & LoPA-Dist-NV & 708.48 & 1470.95 & 15.55 & 0.74 & 619.33 & 1299.25 & 13.16 & 0.85 \\
        \rowcolor{lightgray} \cellcolor{white} \multirow{-2}{*}{D2F-Dream-Base} & LoPA-Dist-Ascend & 1073.86 & 2400.12 & 11.92 & 0.78 & 856.46 & 2751.61 & 9.34 & 0.75 \\

        \midrule
         & LoPA-Dist-NV & 636.55 & 1811.71 & 9.52 & 0.14 & 609.90 & 1407.56 & 11.42 & 0.26 \\
        \rowcolor{lightgray} \cellcolor{white} \multirow{-2}{*}{D2F-Dream-Instruct} & LoPA-Dist-Ascend & 896.21 & 2586.73 & 8.64 & 0.11 & 897.10 & 1868.16 & 9.30 & 0.21 \\
        \bottomrule
    \end{tabular}
    } 
    \caption{\textbf{System performance of LoPA.} The results demonstrate that our system efficiently translates algorithmic parallelism (high TPF) into significant wall-clock acceleration, achieving average throughputs exceeding 1000 tokens/s on the specialized LoPA-Dist-Ascend engine. 
    }
    \label{tab:dream_sys_performance}
\end{table*}

\begin{figure*}[t]
    \centering
    \includegraphics[width=0.9\textwidth]{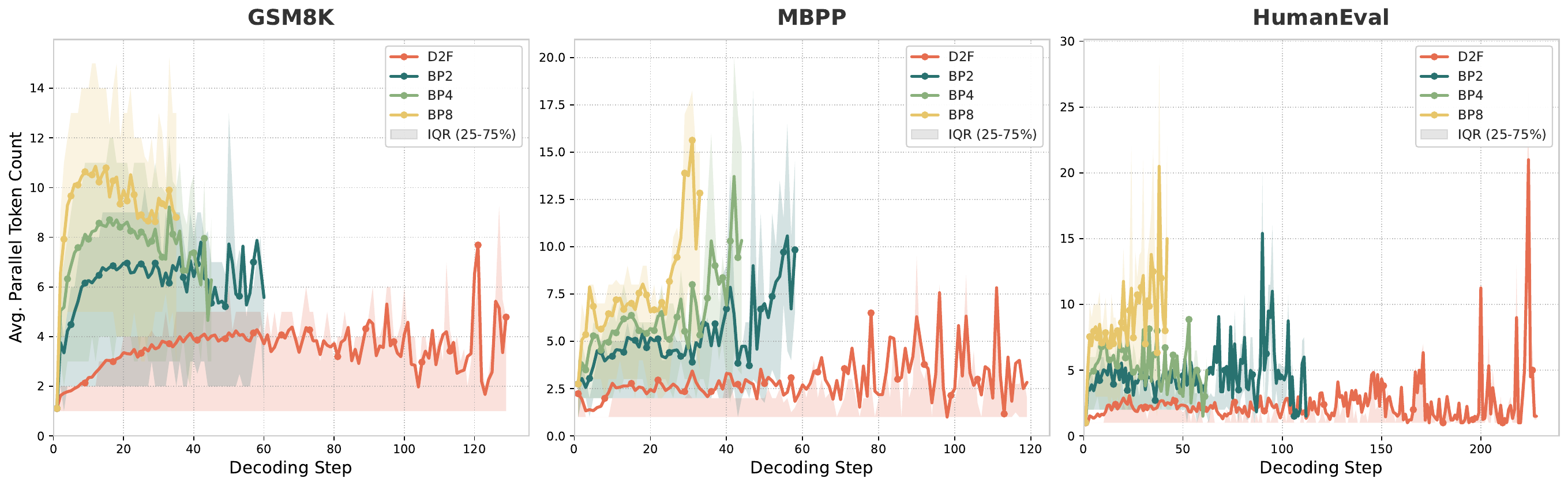} 
    
    \captionof{figure}{\textbf{Scaling analysis of LoPA on D2F-Dream with varying branch counts.} The results illustrate that LoPA effectively scales the TPF of D2F to a peak exceeding 10, thereby significantly reducing the total number of decoding steps.}
    \label{fig:analyse_scaling}
\end{figure*}

\section{Experiments}

\subsection{Experimental Settings}

Our experiments primarily focus on the D2F model~\cite{d2f}, which is the first open-source dLLM whose inference throughput surpasses that of autoregressive (AR) models. Specifically, we integrate LoPA with the Dream-Instruct-7B~\cite{dream} and DiffuCoder-Instruct-7B~\cite{diffucoder} models trained with the D2F objective, denoted as D2F-Dream and D2F-DiffuCoder, respectively. To further verify the generalizability of our method across different architectures, we also integrate LoPA into the vanilla Dream-Instruct-7B model utilizing confidence-based decoding.

Regarding the hardware configuration, we adopt distinct setups for theoretical analysis and system-level evaluation. For the theoretical speedup experiments, we target a distributed setting with 8 NVIDIA A100 GPUs utilizing standard task parallelism. Note that the performance metrics in this phase are projected based on single-GPU benchmarks to derive theoretical upper bounds. For the system-level acceleration experiments, we deploy our method on two distinct high-performance platforms: (1) for the CUDA platform, we utilize 8 NVIDIA H200 GPUs to evaluate performance under Branch Parallelism (BP) configurations of 4 and 8 (BP4 and BP8); (2) for the Ascend platform, we employ a cluster of 8 Ascend 910C NPUs, configured to support a hybrid parallelism strategy combining Tensor Parallelism (TP4) and Branch Parallelism (BP4).

\subsection{Main Results}

\textbf{Baselines.} We compare LoPA against the vanilla dLLM baseline. Additionally, we include Fast-dLLM~\cite{fast-dllm} as a competitive accelerated baseline and the standard D2F model~\cite{d2f} to isolate the performance gains attributable to our lookahead strategy.

\textbf{Benchmarks.} Our evaluation spans diverse domains including mathematical reasoning and code generation. For mathematical reasoning, we employ GSM8K~\cite{gsm8k} and MATH~\cite{math}. For code generation, we utilize HumanEval~\cite{humaneval} and MBPP~\cite{mbpp}. Furthermore, to ensure a rigorous and robust assessment, we include the extended benchmarks HumanEval+ and MBPP+ from EvalPlus~\cite{evalplus}.

\textbf{Scaling D2F Inference.} We analyze the speed-quality trade-off by varying branch count $k$. As detailed in Figure~\ref{fig:scale}, while increasing $k$ enhances parallelism, excessive branching may induce fluctuations by prioritizing future confidence. Consequently, with an optimal $k$, LoPA substantially scales the inference speed. Specifically, it increases the TPF of D2F-Dream to 10.1 on GSM8K while maintaining performance superior to the Dream baseline, and scales D2F-DiffuCoder to 8.3 on HumanEval+ with marginal performance degradation. Tables~\ref{tab:dream_performance} and~\ref{tab:diffucoder_performance} confirm this efficacy. For instance, on MATH, LoPA attains a high TPF of 8.0 while achieving a performance score superior to the Vanilla baseline. This validates that LoPA enables high-parallelism decoding without compromising generation quality.

\textbf{Generalizability Verification.} Beyond D2F-based architectures, LoPA demonstrates strong universality. As shown in Table~\ref{tab:dream_performance}, when applied to the Vanilla Dream utilizing confidence-based decoding., LoPA scales the TPF to 3.4 on Math while maintaining comparable
performance. This evidence validates LoPA as a generalized, plug-and-play scaling solution applicable to broad confidence-driven dLLMs.

\begin{table*}[t]
    \centering
    \setlength{\tabcolsep}{4pt}
    \resizebox{0.95\textwidth}{!}{
    \begin{tabular}{llccccccccc}
        \toprule
        \multirow{2}{*}{\textbf{Model}} & \multirow{2}{*}{\textbf{Sys. Arch.}} & \multirow{2}{*}{\textbf{Settings}} & \multicolumn{4}{c}{\textbf{MBPP 3-shot}} & \multicolumn{4}{c}{\textbf{GSM8K 4-shot}} \\
        \cmidrule(lr){4-7} \cmidrule(lr){8-11}
         & & & \textbf{Avg TPS} & \textbf{Max TPS} & \textbf{Top-10 TPS} & \textbf{Score} & \textbf{Avg TPS} & \textbf{Max TPS} & \textbf{Top-10 TPS} & \textbf{Score} \\
        \midrule
        
        \multirow{18}{*}{D2F-Dream-Base} 
        & \multirow{12}{*}{LoPA-Dist-NV} 
          & S1 & 415.19 & 813.04 & 720.35 & 53.00 & 345.52 & 959.05 & 704.39 & 75.97 \\
        & & S2 & 500.33 & 1185.77 & 874.87 & 53.40 & 402.52 & 913.12 & 842.83 & 73.54 \\
        & & S3 & 550.37 & 1472.41 & 929.72 & 51.20 & 436.22 & 994.82 & 885.27 & 71.19 \\
        & & S4 & 589.22 & 1576.93 & 1006.57 & 47.20 & 475.58 & 1203.61 & 1028.15 & 68.16 \\
        & & S5 & 633.16 & 1408.40 & 963.67 & 46.80 & 516.85 & 1212.65 & 1055.08 & 66.79 \\
        & & S6 & 678.26 & 1615.30 & 1150.65 & 41.80 & 546.72 & 1225.21 & 1121.57 & 64.14 \\
        \cmidrule(lr){3-11}
        & & S7 & 466.27 & 784.33 & 764.52 & 51.80 & 416.91 & 909.82 & 841.95 & 71.27 \\
        & & S8 & 545.90 & 1497.22 & 927.67 & 51.40 & 486.94 & 1176.14 & 959.37 & 68.39 \\
        & & S9 & 588.00 & 1584.28 & 983.09 & 48.60 & 520.70 & 1250.67 & 1056.01 & 68.01 \\
        & & S10 & 637.38 & 1552.56 & 1028.97 & 47.00 & 558.01 & 1115.26 & 1071.66 & 65.05 \\
        & & S11 & 655.45 & 1535.10 & 1059.72 & 43.80 & 592.94 & 1315.93 & 1155.11 & 64.44 \\
        & & S12 & 708.48 & 1470.95 & 1132.78 & 39.80 & 619.33 & 1299.25 & 1201.18 & 60.88 \\
        \cmidrule{2-11} 
        & \multirow{6}{*}{LoPA-Dist-Ascend} 
          & S13 & 615.74 & 2173.7 & 1253.07 & 50.20 & 492.94 & 1337.60 & 1158.18 & 75.06 \\
        & & S14 & 753.78 & 2115.55 & 1397.85 & 50.20 & 589.77 & 1532.99 & 1342.79 & 72.86 \\
        & & S15 & 842.97 & 2470.79 & 1538.16 & 50.00 & 644.34 & 1723.19 & 1476.24 & 70.58 \\
        & & S16 & 923.35 & 2647.12 & 1513.54 & 45.60 & 700.14 & 1756.58 & 1601.93 & 68.69 \\
        & & S17 & 994.88 & 2740.54 & 1739.85 & 43.00 & 754.75 & 2583.76 & 1848.82 & 64.29 \\
        & & S18 & 1073.86 & 2400.12 & 1939.22 & 41.80 & 856.46 & 2751.61 & 2098.72 & 62.55 \\
        \midrule
        
        \multirow{18}{*}{D2F-Dream-Instruct} 
        & \multirow{12}{*}{LoPA-Dist-NV} 
          & S1 & 305.74 & 959.00 & 695.88 & 52.80 & 330.62 & 758.34 & 674.53 & 78.17 \\
        & & S2 & 373.23 & 1302.99 & 877.12 & 51.40 & 402.63 & 961.29 & 804.31 & 74.22 \\
        & & S3 & 451.62 & 1419.09 & 1143.30 & 53.00 & 444.73 & 943.22 & 870.85 & 73.39 \\
        & & S4 & 503.71 & 1779.60 & 1226.72 & 46.60 & 495.93 & 1131.64 & 941.23 & 72.48 \\
        & & S5 & 568.65 & 1660.89 & 1317.38 & 42.00 & 540.76 & 1185.14 & 1033.60 & 68.99 \\
        & & S6 & 615.95 & 1951.86 & 1542.82 & 37.60 & 568.75 & 1352.22 & 1139.06 & 65.88 \\
        \cmidrule(lr){3-11}
        & & S7 & 325.15 & 697.49 & 620.42 & 50.80 & 379.42 & 839.65 & 710.10 & 75.28 \\
        & & S8 & 408.37 & 1182.69 & 866.90 & 51.00 & 449.56 & 934.55 & 838.35 & 75.13 \\
        & & S9 & 465.55 & 1097.40 & 1016.91 & 50.60 & 497.47 & 1172.31 & 946.98 & 74.75 \\
        & & S10 & 544.72 & 1542.99 & 1145.55 & 46.80 & 539.28 & 1147.95 & 1021.96 & 71.34 \\
        & & S11 & 591.57 & 1578.00 & 1204.05 & 42.20 & 580.04 & 1292.18 & 1132.19 & 66.94 \\
        & & S12 & 636.55 & 1811.71 & 1500.59 & 36.00 & 609.90 & 1407.56 & 1159.28 & 65.50 \\
        \cmidrule{2-11} 
        & \multirow{6}{*}{LoPA-Dist-Ascend} 
          & S13 & 412.90 & 911.73 & 911.73 & 50.80 & 515.01 & 1235.84 & 1090.45 & 76.12  \\
        & & S14 & 525.66 & 1546.34 & 1143.37 & 48.40 & 619.58 & 1424.32 & 1310.35 & 75.36  \\
        & & S15 & 625.53 & 1729.78 & 1435.06 & 46.20 & 689.89 & 1644.74 & 1356.36 & 72.63 \\
        & & S16 & 716.19 & 1780.41 & 1558.00 & 43.80 & 770.78 & 1589.69 & 1480.56 & 71.49 \\
        & & S17 & 796.65 & 1798.14 & 1687.69 & 39.80 & 837.21 & 1782.80 & 1517.90 & 67.78 \\
        & & S18 & 896.21 & 2586.73 & 2086.04 & 36.40 & 897.10 & 1868.16 & 1642.72 & 66.87 \\
    
        \bottomrule
    \end{tabular}
    }
    \caption{\textbf{Performance ablation study of D2F-Dream models on different platforms, corresponding to settings S1-S18.} The results illustrate the trade-off between inference throughput and generation quality across varying branch configurations and system backends.}
    \label{tab:dream_sys_performance_ablation}
\end{table*}

\textbf{System Throughput.} To fully unleash the parallelism enabled by LoPA across diverse hardware backends, we evaluate the wall-clock performance using our co-designed Branch Parallel (BP) inference system, as illustrated in Figure~\ref{fig:lopa_sys}. Under multi-device deployment, the system exhibits near-linear scalability with respect to the branch count, effectively translating high TPF into tangible throughput gains. As shown in Figure~\ref{fig:tps_comparison} and Table~\ref{tab:dream_sys_performance}, the system achieves a single-sample throughput of \textbf{1073.86} tokens per second, where both the D2F baseline and D2F + LoPA employ identical, slightly lowered decoding thresholds to prioritize and maximize inference speed.

The comprehensive evaluation results are summarized in Table~\ref{tab:dream_sys_performance}, where the optimal configurations selected for peak performance are TP1+BP8 for LoPA-Dist-NV and TP4+BP4 for LoPA-Dist-Ascend. It is important to note that this table presents the peak system throughput achieved under the optimal hyperparameter configurations mentioned above. For a granular analysis of performance across different settings (e.g., varying decoding thresholds and parallel strategies), please refer to the detailed ablation study provided in Table~\ref{tab:dream_sys_performance_ablation}.

It is noteworthy that we provide a cross-platform implementation to demonstrate the versatility of our system. While both implementations adhere to the unified architectural design shown in Figure~\ref{fig:lopa_sys}, they employ distinct backend strategies: the \textbf{CUDA} results correspond to a naive distributed implementation, whereas the \textbf{Ascend} results are derived from our specialized inference engine, which incorporates PagedAttention-like~\cite{vllm} mechanisms adapted for deployment on Ascend devices. In addition to the standard scores, we detail the system performance metrics including the Average TPS, which reflects the sustained generation speed; the Max TPS, demonstrating the peak throughput capability; TPF, which quantifies the efficiency of parallel decoding; and Latency, which provides a direct measure of wall-clock inference speed. 

\textbf{Scaling Analysis.} As shown in Figure~\ref{fig:analyse_scaling}, we conduct a comprehensive scaling analysis of LoPA on D2F-Dream with varying branch counts across multiple benchmarks. 
The results illustrate that LoPA effectively scales the TPF of D2F to a peak exceeding 10, thereby significantly reducing the total number of decoding steps. 
We further observe that math tasks like GSM8K exhibit high parallelism in the middle stages of generation, whereas code tasks such as MBPP and HumanEval show higher parallelism in the later stages.

\section{Conclusion}

In this paper, we propose \textbf{Lookahead Parallel Decoding (LoPA)}, a training-free algorithm designed to break the parallelism bottleneck in dLLM inference by identifying the optimal Token Filling Order (TFO). 
By concurrently exploring candidate branches to maximize future confidence, LoPA significantly scales decoding efficiency, boosting the Tokens Per Forward pass (TPF) of D2F-Dream to \textbf{10.1} on GSM8K and D2F-DiffuCoder to \textbf{8.3} on HumanEval+ while preserving generation quality. 
Furthermore, we facilitate these gains with a specialized \textit{Branch Parallel (BP)} inference system that ensures substantial wall-clock speedups, establishing LoPA as a robust solution for efficient non-autoregressive sequence generation.

\nocite{langley00}

\bibliography{reference}
\bibliographystyle{icml2026}

\newpage
\appendix
\onecolumn

\section{Hardware Environment: Huawei Ascend 910C}

Diffusion models typically involve iterative denoising steps, leading to repetitive kernel launches. We utilize the graph compilation capabilities of the CANN software stack to fuse element-wise operations (such as bias-add and activation functions) and optimize the computation graph. This reduces the kernel launch overhead significantly during the multi-step diffusion sampling process.

The Ascend inference engine is deployed on the Huawei Ascend 910C AI processor, a next-generation Neural Processing Unit (NPU) designed for large-scale AI training and inference. The Ascend 910C is built upon the advanced Da Vinci architecture, featuring high-performance Cube Units for matrix operations and Vector Units for general-purpose calculations.

Compared to its predecessors, the 910C offers significant improvements in compute density and inter-chip interconnect bandwidth, making it highly suitable for the memory-intensive nature of Diffusion Language Models.

The platform utilizes the Computational Architecture for Neural Networks (CANN), which acts as the bridge between the upper-level deep learning frameworks and the underlying hardware resources, enabling efficient operator mapping and memory management.

\section{Detailed Hyperparameter Configurations}

In this section, we present the specific hyperparameter settings used for the D2F-Dream LoPA and D2F-DiffuCoder LoPA experiments. Table \ref{tab:hyperparams} details the prompt shot counts, D2F parameters (including Block size, $\tau_{add}$, $\tau_{act}$, and $\tau_{conf}$), and the number of LoPA branches for each dataset. These configurations correspond to the evaluation results reported in Table \ref{tab:dream_performance} and Table \ref{tab:diffucoder_performance}.

\begin{table}[h]
    \centering
    \begin{tabular}{lcccccc}
        \toprule
        \textbf{Dataset} & \textbf{Shots} & \textbf{Block Size} & $\bm{\tau_{add}}$ & $\bm{\tau_{act}}$ & $\bm{\tau_{conf}}$ & \textbf{LoPA Branches} \\
        \midrule
        \multicolumn{7}{l}{\textit{\textbf{D2F-Dream LoPA}}} \\ 
        GSM8K     & 4-Shot & 32 & 0.1 & 0.95 & 0.90 & 14 \\
        MATH      & 4-Shot & 16 & 0.1 & 0.95 & 0.90 & 7  \\
        HumanEval & 0-Shot & 32 & 0.3 & 0.95 & 0.95 & 7  \\
        MBPP      & 3-Shot & 32 & 0.3 & 0.95 & 0.95 & 7  \\
        \midrule
        \multicolumn{7}{l}{\textit{\textbf{D2F-DiffuCoder LoPA}}} \\ 
        HumanEval+ & 0-Shot & 32 & 0.3 & 0.95 & 0.95 & 6  \\
        MBPP+      & 0-Shot & 32 & 0.3 & 0.95 & 0.90 & 14 \\
        \bottomrule
    \end{tabular}
    
    \caption{Hyperparameter settings for D2F-Dream LoPA and D2F-DiffuCoder LoPA. The listed parameters correspond to the main results in Table \ref{tab:dream_performance} and Table \ref{tab:diffucoder_performance}.}
    \label{tab:hyperparams}
\end{table}

\begin{table}[t]
    \centering
    \begin{tabular}{clcccccccc}
        \toprule
        \multirow{2}{*}{\textbf{Settings}} & \multirow{2}{*}{\textbf{\textbf{Sys. Arch.}}} & \multirow{2}{*}{\textbf{Precision}} & \textbf{TP} & \textbf{BP} & \textbf{Block} & \textbf{Seq.} & \multirow{2}{*}{$\tau_{\text{add}}$} & \multirow{2}{*}{$\tau_{\text{act}}$} & \multirow{2}{*}{$\tau_{\text{conf}}$} \\
         & & & \textbf{Size} & \textbf{Size} & \textbf{Size} & \textbf{Length} & & & \\
        \midrule
        
        S1  & \multirow{12}{*}{LoPA-Dist-NV} 
              & BF16 & 1 & 4 & 32 & 512 & 0.1 & 0.95 & 0.95  \\
        S2  & & BF16 & 1 & 4 & 32 & 512 & 0.1 & 0.9 & 0.9  \\
        S3  & & BF16 & 1 & 4 & 32 & 512 & 0.1 & 0.85 & 0.85  \\
        S4  & & BF16 & 1 & 4 & 32 & 512 & 0.1 & 0.8 & 0.8  \\
        S5  & & BF16 & 1 & 4 & 32 & 512 & 0.1 & 0.75 & 0.75  \\
        S6  & & BF16 & 1 & 4 & 32 & 512 & 0.1 & 0.7 & 0.7  \\
        S7  & & BF16 & 1 & 8 & 32 & 512 & 0.1 & 0.95 & 0.95  \\
        S8  & & BF16 & 1 & 8 & 32 & 512 & 0.1 & 0.9 & 0.9  \\
        S9  & & BF16 & 1 & 8 & 32 & 512 & 0.1 & 0.85 & 0.85  \\
        S10  & & BF16 & 1 & 8 & 32 & 512 & 0.1 & 0.8 & 0.8  \\
        S11  & & BF16 & 1 & 8 & 32 & 512 & 0.1 & 0.75 & 0.75  \\
        S12  & & BF16 & 1 & 8 & 32 & 512 & 0.1 & 0.7 & 0.7  \\
        
        \midrule
        
        S13  & \multirow{6}{*}{LoPA-Dist-Ascend} 
              & W8A8 & 4 & 4 & 32 & 512 & 0.1 & 0.95 & 0.95  \\
        S14  & & W8A8 & 4 & 4 & 32 & 512 & 0.1 & 0.9 & 0.9  \\
        S15  & & W8A8 & 4 & 4 & 32 & 512 & 0.1 & 0.85 & 0.85  \\
        S16  & & W8A8 & 4 & 4 & 32 & 512 & 0.1 & 0.8 & 0.8  \\
        S17  & & W8A8 & 4 & 4 & 32 & 512 & 0.1 & 0.75 & 0.75  \\
        S18  & & W8A8 & 4 & 4 & 32 & 512 & 0.1 & 0.7 & 0.7  \\
        \bottomrule
    \end{tabular}
    \caption{\textbf{Hyperparameter configurations for each setting employed in the performance ablation study.} \textbf{TP}: Tensor Parallel, \textbf{BP}: Branch Parallel.}
    \label{tab:hyperparameter_settings}
\end{table}
\section{Detailed Baseline Results}

To provide a comprehensive performance assessment, we compare our method against representative state-of-the-art models from both the Diffusion LLM and Autoregressive (AR) LLM families. Specifically, we include SDAR~\cite{sdar} as a strong dLLM baseline known for its high generation quality, and Qwen2.5-7B-Instruct as a standard benchmark for AR inference speed and accuracy. Table~\ref{tab:baseline_results} details these comparisons, highlighting the trade-offs between throughput (TPS), parallelism (TPF), and 
generation scores.

\begin{table*}[t]
    \centering
    \setlength{\tabcolsep}{4pt}
    \resizebox{0.85\textwidth}{!}{%
    \begin{tabular}{lccccccccc}
        \toprule
        \multirow{2}{*}{\textbf{Model}} & \multicolumn{4}{c}{\textbf{MBPP}} & \multicolumn{4}{c}{\textbf{GSM8K}} \\
        \cmidrule(lr){2-5} \cmidrule(lr){6-9}
        & \textbf{Avg TPS} & \textbf{TPF} & \textbf{Latency} & \textbf{Score} & \textbf{Avg TPS} & \textbf{TPF} & \textbf{Latency} & \textbf{Score} \\
        \midrule
        D2F-Dream-7B-Base (CUDA) & 327.69 & 5.64 & 1.91 & 45.00 & 224.33 & 3.90 & 2.68 & 64.90 \\
        D2F-Dream-7B-Instruct (CUDA) & 206.37 & 3.13 & 0.54 & 45.00 & 247.90 & 4.01 & 0.89 & 69.07 \\
        Qwen3-8B (SGLang) & 317.41 & 1.00 & 0.84 & 78.92 & 317.31 & 1.00 & 0.49 & 93.63 \\
        SDAR-4B-Chat & -- & 1.50 & -- & 52.0 & -- & 2.00 & -- & 88.90 \\
        
        SDAR-8B-Chat (LMDeploy) & 234.90 & 1.60 & 0.25 & 72.00 & 273.31 & 2.10 & 0.72 & 91.30 \\
        LLaDA2.0-flash (SGLang) & 433.81 & 2.70 & 0.21 & 88.30 & 307.92 & 5.30 & 0.64 & 96.06 \\
        \bottomrule
    \end{tabular}
    }
    \caption{\textbf{Performance comparison of baseline results on multiple benchmarks across multiple platforms.} In this configuration, D2F hyperparameters are specifically aligned with the peak throughput capabilities of the current LoPA system, where decoding thresholds are adjusted to ensure maximum inference speed. We incorporate SDAR-4B-Chat and Qwen3-8B to benchmark against SOTA dLLMs and AR models. TPS denotes Tokens Per Second. The Ascend engine demonstrates the effectiveness of our system design compared to the naive CUDA implementation.}
    \label{tab:baseline_results}
\end{table*}


\end{document}